\documentclass[letterpaper]{article} 
\usepackage{aaai2026}  
\usepackage{times}  
\usepackage{helvet}  
\usepackage{courier}  
\usepackage[hyphens]{url}  
\usepackage{graphicx} 
\usepackage{natbib}  
\usepackage{caption} 
\usepackage{algorithm}
\usepackage{algorithmic}
\usepackage{amsmath}
\usepackage{array} 
\usepackage{tabularx}
\usepackage{booktabs}
\usepackage{multirow}

\usepackage{newfloat}
\usepackage{listings}
\DeclareCaptionStyle{ruled}{labelfont=normalfont,labelsep=colon,strut=off} 
\floatstyle{ruled}
\newfloat{listing}{tb}{lst}{}
\floatname{listing}{Listing}
\title{Debate over Mixed-knowledge: A Robust Multi-Agent Reasoning Framework for Incomplete Knowledge Graph Question Answering}

\newcommand{\shortname}{DoM}

\author{
    Jilong Liu\textsuperscript{\rm 1} , 
    Pengyang Shao\footnotemark[1] \textsuperscript{\rm 2}, 
    Wei Qin\textsuperscript{\rm 1} , 
    Fei Liu\textsuperscript{\rm 1 3} , 
    Yonghui Yang\textsuperscript{\rm 2} , 
    Richang Hong\thanks{Corresponding authors} \textsuperscript{\rm 1}
}
\affiliations{
    \textsuperscript{\rm 1}Hefei University of Technology\\
    \textsuperscript{\rm 2}National University of Singapore\\
    \textsuperscript{\rm 3}The Key Laboratory of Knowledge Engineering with Big Data (the Ministry of Education of China)\\
    \{liujilong0116, shaopymark, qinwei.hfut\}@gmail.com,feiliu@mail.hfut.edu.cn, \{yyh.hfut, hongrc.hfut\}@gmail.com

}

\usepackage{bibentry}

\begin{document}

\maketitle

\begin{abstract}




Knowledge Graph Question Answering (KGQA) aims to improve factual accuracy by leveraging structured knowledge. However, real-world Knowledge Graphs (KGs) are often incomplete, leading to the problem of Incomplete KGQA (IKGQA). A common solution is to incorporate external data to fill knowledge gaps, but existing methods lack the capacity to adaptively and contextually fuse multiple sources, failing to fully exploit their complementary strengths.
To this end, we propose Debate over Mixed-knowledge (DoM), a novel framework that enables dynamic integration of structured and unstructured knowledge for IKGQA. Built upon the Multi-Agent Debate paradigm, DoM assigns specialized agents to perform inference over knowledge graphs and external texts separately, and coordinates their outputs through iterative interaction. It decomposes the input question into sub-questions, retrieves evidence via dual agents (KG and Retrieval-Augmented Generation, RAG), and employs a judge agent to evaluate and aggregate intermediate answers. This collaboration exploits knowledge complementarity and enhances robustness to KG incompleteness.
In addition, existing IKGQA datasets simulate incompleteness by randomly removing triples, failing to capture the irregular and unpredictable nature of real-world knowledge incompleteness. To address this, we introduce a new dataset, Incomplete Knowledge Graph WebQuestions, constructed by leveraging real-world knowledge updates. These updates reflect knowledge beyond the static scope of KGs, yielding a more realistic and challenging benchmark. Through extensive experiments, we show that DoM consistently outperforms state-of-the-art baselines.
\end{abstract}


\section{Introduction}



Recent advances in Knowledge Graph-based Question Answering (KGQA) have shown that augmenting larges language models (LLMs) with structured, semantically rich knowledge graphs (KGs) can improve the factual reliability of model outputs~\cite{luo2024reasoning, chen2024plan, tan2025paths, ma2025debate}. These methods typically retrieve KG subgraphs relevant to the input query and feed them into the LLM in a multi-step manner, thereby improving answer quality~\cite{sun2023think}. Although effective, these methods tend to rely on the completeness of the underlying knowledge graphs—conditions that are difficult to satisfy in practice due to the high cost of KG construction and maintenance~\cite{hur2021survey}. Previous studies have recognized the challenge of KG incompleteness~\cite{min2013distant, ren2020query2box, pflueger2022gnnq}, which has led to the emergence of \textbf{Incomplete Knowledge Graph Question Answering (IKGQA)} as a distinct research task. 

To address the IKGQA issue, existing approaches can be broadly categorized to: (i) KG-internal completion, and (ii) external information augmentation. For category (i), KG-internal completion methods typically aim to predict missing links by learning embedding representations of entities and relations, modeling relational patterns among existing triples within the KG~\cite{saxena2022sequence, guo2023knowledge}. However, such methods inherently rely on existing KG structure and thus struggle to capture changes in the external world, as real-world KG incompleteness often arises from evolving events. Category (ii) includes methods that mitigate KG incompleteness by incorporating knowledge sources beyond the KG itself. some approaches leverage external textual corpora, such as Wikipedia, to construct question-specific subgraphs and enrich the KG with supplementary contextual information~\cite{sun2019pullnet, lv2020graph}. Others treat the parametric knowledge embedded in LLMs as an auxiliary information source, using it to infer missing triples~\cite{xu2024generate}. While these methods demonstrate the value of incorporating external information to mitigate KG incompleteness, they still exhibit notable limitations. Approaches that integrate structured and unstructured data typically rely on training models with substantial amounts of aligned data, making them sensitive to the quality and coverage of training resources. On the other hand, methods that rely exclusively on the parametric knowledge encoded in LLMs may yield incorrect inferences when the missing KG facts fall beyond the coverage of the LLMs' pre-training. These limitations raise a central question: How to leverage inference capabilities of LLM to better integrate mixed knowledge for the IKGQA issue?

\begin{figure}[t]
    \centering
    \includegraphics[width=\linewidth]{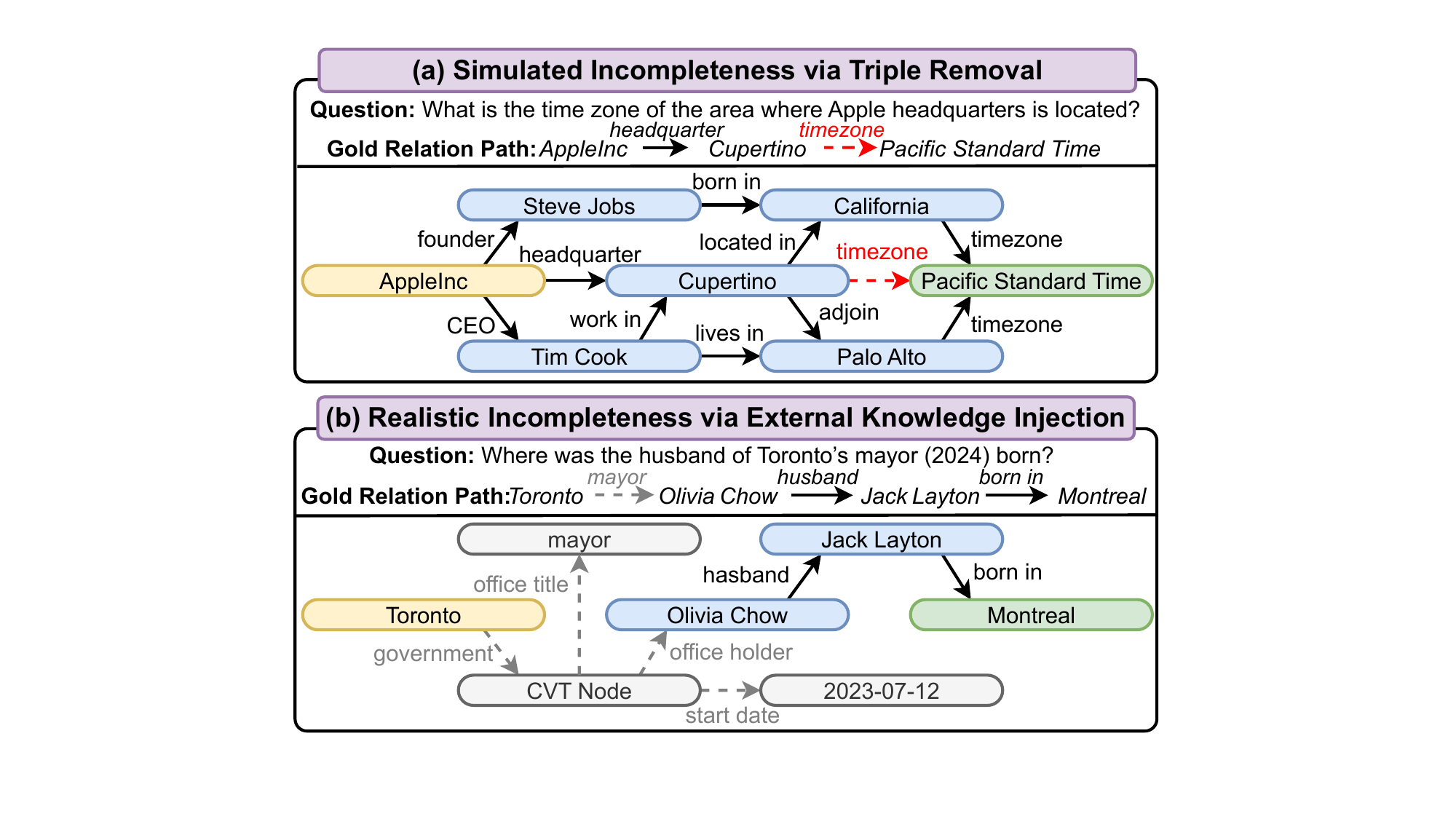}
    \caption{Illustration of two strategies for constructing incomplete KG scenarios. (a) Simulated incompleteness by removing triples from the KG; red dashed arrows denote the deleted facts. (b) Realistic incompleteness by incorporating external knowledge; gray dashed arrows indicate newly introduced information. CVT (Compound Value Type) nodes are used to represent multi-entity relations in Freebase.}
    \label{fig:IKG_Construction}
\end{figure}

To this end, we propose Debate over Mixed-knowledge (DoM). Given the need to effectively integrate external and KG-based knowledge, we adopt the Multi-Agent Debate (MAD) framework, where independent agents specialize in performing inference over different knowledge sources and collaboratively refine their outputs through interaction and alignment. Specifically, to effectively combine MAD with IKGQA, we introduce a three-stage framework : (1) Initialization: to prepare for effective coordination of mixed-knowledge, we decompose the input question into semantically coherent sub-questions. These sub-questions are dynamically updated throughout the inference process. (2) Sub-question Inference: to integrate heterogeneous knowledge, we design dual retrieval agents, including a KG Agent for structured KG and a Retrieval-Augmented Generation (RAG) Agent for unstructured external knowledge. Each agent independently retrieves candidate evidence, and a Judge Agent evaluates and integrates their outputs to produce an intermediate answer and update the inference plan. This design enables mutual correction and complementarity between knowledge sources. (3) Final Answer Generation: to ensure global consistency, DoM prompts an LLM to consolidate intermediate results into a final answer. However, existing IKGQA datasets typically simulate incompleteness by randomly removing gold-path triples. This synthetic pattern fails to capture the irregular and evolving nature of real-world incomplete KG scenarios. To address this limitation, we construct a new dataset, Incomplete Knowledge Graph WebQuestions (IKGWQ), by revisiting existing QA benchmarks (CWQ and WebQSP) and regenerating question-answer pairs using up-to-date knowledge. This design naturally introduces both missing triples and missing entities, thereby capturing the evolving nature of real-world KG incompleteness.
Finally, extensive experiments demonstrate the effectiveness of our proposed \shortname. Our main contributions are summarized as follows:


\begin{enumerate}
    \item To better integrate multi-source knowledge to address the IKGQA challenge, we propose a multi-agent debate framework, Debate over Mixed-knowledge (DoM), which enables dynamic and complementary inference from incomplete KGs and external knowledge.
    \item We construct a new dataset, IKGWQ, addressing the limitations of existing IKGQA datasets by introducing real-world factual updates, better reflecting the irregularity and unpredictability of KG incompleteness.
    \item DoM achieves consistent gains over strong baselines, with up to 13.6\% relative improvement in Hits@1 on existing IKGQA datasets and 70.7\% on IKGWQ.
\end{enumerate}

\section{Data Description}

Existing IKGQA datasets often simulate incompleteness by removing triples from gold relation paths, as illustrated in Figure~\ref{fig:IKG_Construction}(a). To better capture real-world KG dynamics, we construct IKGWQ by revisiting samples from CWQ and WebQSP and rebuilding corresponding question-answer using up-to-date knowledge, as shown in Figure~\ref{fig:IKG_Construction}(b).

\begin{figure}[t]
    \centering
    \includegraphics[width=\linewidth]{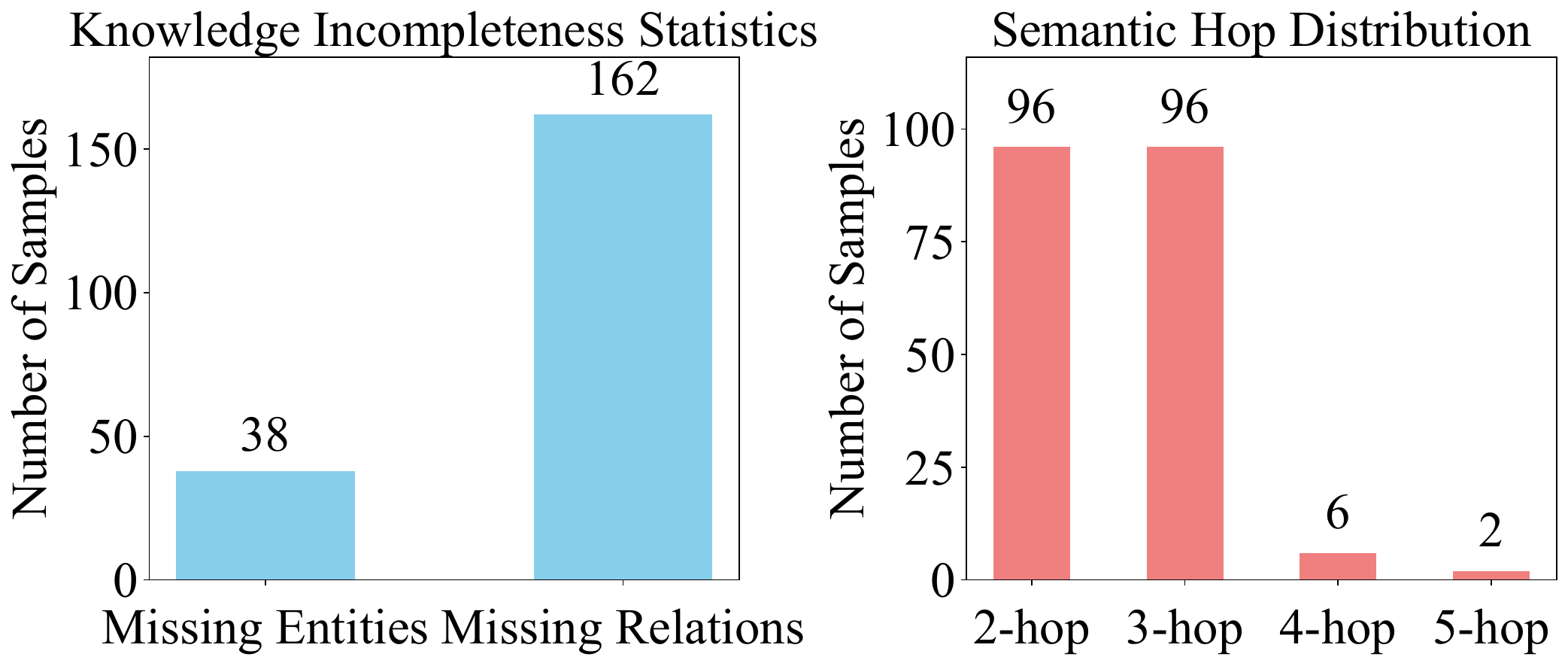}
    \caption{Statistics of knowledge incompleteness and semantic hop distribution in the IKGWQ dataset. The left subfigure shows the number of samples with missing entities and missing relations. The right subfigure presents the distribution of semantic inference hops.}
    \label{fig:dataset_analysis}
\end{figure}

Specifically, we revisit each selected sample by retrieving up-to-date information for its topic entity from reliable sources and constructing question-answer pairs that reflect facts missing or outdated in the original Freebase~\cite{bollacker2008freebase} KG.  First, we extract topic entities from the original datasets and retrieve their updated descriptions via the Wikipedia API. We then guide an LLM to generate question-answer pairs based on knowledge extracted from these texts—focusing particularly on facts beyond the scope of the original KG. Finally, we conduct manual filtering, question rewriting, and human annotation to ensure the quality and correctness of the resulting dataset. This pipeline naturally introduces both missing triples and missing entities, offering a more realistic simulation of the dynamic and uncertain nature of KG incompleteness in real-world scenarios.

To better understand the characteristics and challenges posed by the IKGWQ dataset, we conduct a detailed analysis in two key dimensions: knowledge incompleteness and inference complexity. The results are summarized in Figure~\ref{fig:dataset_analysis}. The dataset comprises 200 samples, including 38 instances of missing entities and 162 instances of missing relations. Note that missing-entity cases inherently subsume relation incompleteness, whereas missing-relation cases do not involve any entity omission. In terms of inference complexity, a substantial portion of questions require multi-hop inference: 48\% of the samples involve 3-hop inference, with some extending up to 5-hop. Note that hop count is defined based on the number of semantic inference steps needed to answer a question, excluding auxiliary nodes such as Compound Value Type nodes in Freebase. The actual KG traversal steps may be higher due to these intermediate structures.

\section{Preliminary}

\subsection{Incomplete Knowledge Graph Question Answering}
IKGQA is a generalization of the standard KGQA task. In KGQA, the goal is to predict the correct answer entities $A_q \subseteq E$ for a given natural language question $q$, based on a complete knowledge graph $G = (E, R, T)$. Here, $E$ denotes the set of entities, $R$ denotes the set of relations, and $T = { (e_h, r, e_t) \mid e_h, e_t \in E, r \in R }$ represents the set of factual triples. Each triple consists of a head entity $e_h$, a relation $r$, and a tail entity $e_t$. This task assumes that all topic entities $T_q \subseteq E$ and the necessary relation paths are present in $G$. Formally, KGQA can be defined as a function ${f: (q, G) \mapsto A_q}$.

In real-world applications, however, KGs are often incomplete due to limitations in construction and maintenance. To address this, IKGQA relaxes the requirement for full KG coverage by enabling the reasoning process to leverage external knowledge—either retrieved from textual sources or derived from the internal knowledge embedded in LLMs. Following GoG~\cite{xu2024generate}, IKGQA can be formulated as ${f: (q, G, \mathcal{R}) \mapsto A_q}$, where $\mathcal{R}$ denotes external knowledge that supplements the incomplete KG $G$ during the inference process.

\subsection{Notation for LLM Modules}  
We denote the outputs of task-specific LLM calls using the notation $\text{LLM}_{\text{role}}(\cdot)$, where the subscript indicates the role or function performed by the LLM in the inference process. For example, $\text{LLM}_{\text{plan}}$ represents the planning module for sub-question decomposition.

\begin{figure*}[t]
    \centering
    \includegraphics[width=\linewidth]{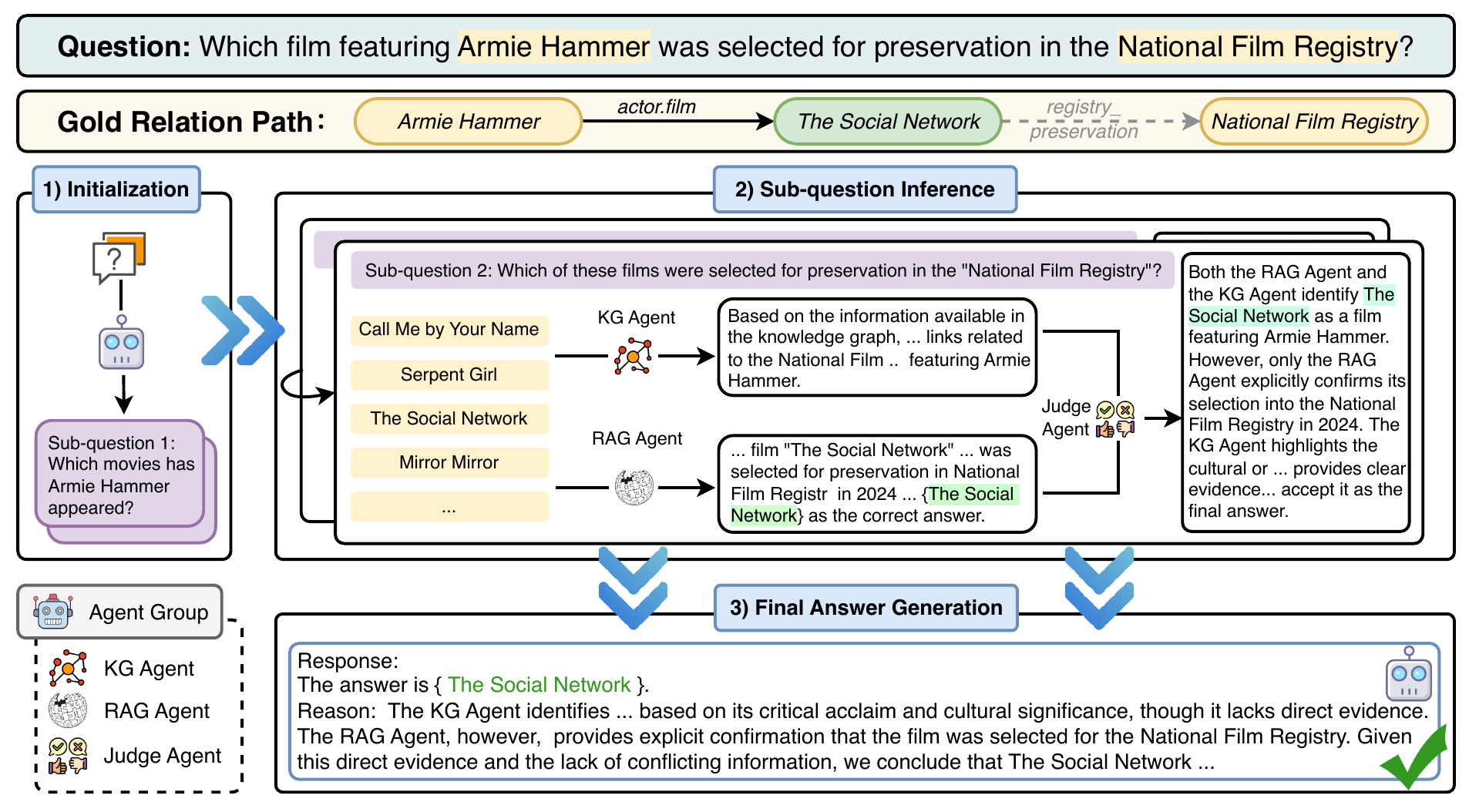}
    \caption{Overview of the DoM framework. DoM first decomposes the input question into sub-questions. For each sub-question, the KG Agent and RAG Agent independently infer over structured and unstructured knowledge, and the Judge Agent integrates their outputs through iterative debate. This interaction continues until sufficient evidence is gathered for final answer generation.}
    \label{fig:overview}
\end{figure*}

\section{Debate over Mixed-knowledge (DoM)} 

The workflow of DoM is showed in Figure~\ref{fig:overview}. At the inference stage, DoM adopts the MAD framework to enhance the utilization of mixed knowledge in IKG scenarios. It constructs a KG Agent and a RAG Agent to independently infer from structured KGs and unstructured external data. Their outputs are then aligned and integrated by a Judge Agent. This pipeline consists of three phases: \textbf{Initialization}, \textbf{Sub-question Inference}, and \textbf{Final Answer Generation}.

\subsection{Initialization}

Given a natural language question $q$, we prompt the LLM to decompose it into a sequence of semantically meaningful sub-questions, forming an ordered list $Q$:
\begin{equation}
Q = \text{LLM}_{plan}(q) = \{q_1, q_2, \dots, q_n\},
\end{equation}
Each sub-question $q_i$ is treated as an intermediate goal and will be addressed sequentially in later stages. This decomposition transforms a complex question into a controllable step-by-step inference trajectory, serving as the structural foundation for the subsequent multi-agent debate. We also initialize a system memory $M$ to store intermediate inference results across sub-questions; its structure will be detailed in the final answer generation phase.

\subsection{Sub-question Inference}
After initialization, DoM enters the inference phase, iteratively integrating structured and unstructured knowledge to solve sub-questions. Each iteration consists of three main steps: \textbf{knowledge graph inference}, \textbf{external knowledge inference}, and \textbf{debate-based integration}.

Formally, at the $i$-th iteration, the current inference context is centered around a set of topic entities, denoted as $E_{i}^{\text{topic}} = \{\hat{e}_{i}^1, \hat{e}_{i}^2, \dots, \hat{e}_{i}^m\}$. Based on these entities, the system retrieves:

\begin{itemize}
    \item A set of KG relation paths $P_i = \{T_i^1, T_i^2, \dots, T_i^m\}$, where each path $T_i^j = \{t_i^{j,1}, t_i^{j,2}, \dots, t_i^{j,l}\}$ is a sequence of triples starting from entity $\hat{e}_i^j$. Each triple $t_i^{j,k} = (h, r, t)$ denotes a factual relation in the KG.

    \item A set of external textual evidence chunks $C_i = \{C_i^1, C_i^2, \dots, C_i^o\}$ retrieved by the RAG Agent.
\end{itemize}

These two sources of knowledge are independently processed by the KG Agent and the RAG Agent. Each agent infers from its respective evidence and proposes a candidate answer to $q_i$. The Judge Agent integrates these outputs and determines the result, based on which the system updates the next sub-question $q_{i+1}$.

\subsubsection{Knowledge Graph Inference}

This process focuses on answering the sub-question $q_i$ using structured knowledge, facilitated by the \textbf{KG Agent}. This KG-based processing pipeline consists of two phases: \textbf{entity linking} and \textbf{iterative exploration}. Although $E_i^{\text{topic}}$ may contain multiple entities, we select a representative $\hat{e}_i$ to illustrate these two steps.

\textbf{Entity Linking} In each iteration (except the first), the topic entities are derived from the previous sub-question’s answer, which may originate from KG, external sources, or the LLM's internal knowledge. Since non-KG-originated entities may not align with the KG, we perform entity linking to map them to corresponding machine identifiers.

This process involves two steps: (1) retrieving top candidate KG entities via embedding-based name similarity search, and (2) prompting the LLM to select the most suitable entity based on relational context in KG.

Formally, for the ungrounded entity mention $\hat{e}_i$, we retrieve the $top_k$ KG entities $E_i'$ whose names are most similar to $\hat{e}_i$ in the embedding space, i.e., $E_i' = top_k(\text{sim}(E(\hat{e}_i), E(G)))$, where $E(\cdot)$ denotes the entity name encoder, $\text{sim}(\cdot, \cdot)$ denotes embedding similarity. Then, for each candidate entity in $E_i'$, we retrieve its associated KG description. The final linked entity $e_i$ is selected by the LLM:
\begin{equation}
e_i = \text{LLM}_{entity\_select}(\hat{e}_i, E_i').
\end{equation}
If all candidate entities are deemed unsuitable, we assume the target entity is missing from the KG. In this case, KG exploration is skipped for this sub-question, and the LLM resorts to internal CoT inference, as shown in Equation~\ref{eq:kg_cot_reasoning}.

\textbf{Iterative KG Exploration} Once the entity $e_i$ is linked to the KG, we initialize the KG inference process by setting $e_i^1 = e_i$, and perform an iterative search for supporting evidence. At the $w$-th KG inference step for $q_i$, we first retrieve all 1-hop outbound relations of the current entity $e_i^w$—including both outbound and inbound edges (i.e., inverse relations)—via SPARQL, denoted as $S_r(e_i^w)$. The LLM then selects the $top_k$ relations most relevant to the current sub-question $q_i$:
\begin{equation}
R_i^{w} = \text{LLM}_{relation\_select}(S_r(e_i^{w}), q_i),
\end{equation}
For each selected relation, we extract the associated triples via $S_t(e_i^w, R_i^w)$, and incrementally update the candidate evidence set as $P_i \leftarrow P_i \cup T_i^w$. The union of retrieved triples across iterations constitutes the KG relation path for sub-question $q_i$. The evidence set $P_i$ is initialized as an empty set and progressively expanded over iterations.

The LLM determines whether the current evidence set $P_i$ is sufficient to answer the sub-question $q_i$. If so, it directly produces the answer $\hat{a}_i^{kg}$; otherwise, it selects a new entity from $P_i$ to update the topic entity $e_i^{w+1}$ for the next iteration:
\begin{equation}
\text{LLM}_{kg\_inference}(P_i, q_i, M) =
\begin{cases}
\hat{a}_i^{kg}, & \text{sufficient;} \\
e_i^{w+1}, & \text{otherwise.}
\end{cases}
\end{equation}
The process continues iteratively until an answer is generated or a maximum of $W$ steps is reached. If no answer is derived within $W$ steps, or if entity linking fails, the LLM resorts to CoT inference:
\begin{equation}
\hat{a}_i^{kg}=\text{LLM}_{CoT}(q_i, M), w=W \text{ or } \hat{e}_i \text{ missing in KG}.
\label{eq:kg_cot_reasoning}
\end{equation}
\subsubsection{External Knowledge Inference}
This process aims to answer the sub-question $q_i$ using external unstructured knowledge, facilitated by the \textbf{RAG Agent}. To ensure factual reliability, we adopt Wikipedia as the external knowledge source.

Given the topic entity $\hat{e}_i$, we retrieve the top-$k$ most relevant Wikipedia articles, which are then segmented into text chunks to construct the candidate evidence set $C_i$. We then compute relevance scores between each chunk $c \in C_i$ and the sub-question $q_i$ using an embedding model. The top-$k$ scored chunks $C_i' = \{c_i^1, \dots, c_i^k\}$ are selected as the contextual input for LLM-based inference.

Similar to the KG Agent, if the retrieved evidence is insufficient, the RAG Agent resorts to the LLM's internal knowledge via CoT inference to preserve the continuity of the inference process.
We formalize the RAG Agent’s inference process as follows: given the candidate evidence set $C_i$, the top-$k$ relevant chunks are selected as $C_i' = top_k(\text{sim}(E(C_i), E(q_i)))$, 
\begin{equation}
\hat{a}_i^{rag}=
\begin{cases}
\text{LLM}_{rag\_inference}(C_i',q_i, M), & \text{sufficient;} \\
\text{LLM}_{CoT}(q_i, M), & \text{otherwise.}
\end{cases}
\end{equation}
\subsubsection{Debate-based Integration}
For each sub-question $q_i$, the KG Agent and RAG Agent independently generate answers accompanied by their respective inference chains. The \textbf{Judge Agent} then serves as an arbiter to evaluate, compare, and integrate these outputs, producing the consolidated sub-answer $a_i$.

A new iteration is triggered either when sub-questions remain, or when existing information is deemed insufficient to answer the original query. The current sub-answer $a_i$ is used to update the topic entity set. If the next sub-question $q_{i+1}$ exists, it is revised based on the context; otherwise, a new sub-question is adaptively generated. This step is defined as:

\begin{equation}
a_i,q_{i+1}^{new} = \text{LLM}_{judge}(q,q_i,q_{i+1},\hat{a}_i^{kg}, \hat{a}_i^{rag}, M).
\end{equation}

This mechanism enables a dynamic and adaptive inference loop, guided by accumulated evidence. To ensure termination, the number of sub-question iterations is capped by a predefined threshold $I$.

\subsection{Final Answer Generation}
After all sub-questions have been processed or the maximum number of inference iterations $I$ has been reached, the system proceeds to generate the final answer. This step is based on the accumulated inference memory $M$, which records intermediate results from each sub-question in the form $M=[[q_1,\hat{a}_1^{kg},\hat{a}_1^{rag},a_1],...]$. The final answer $a_{final}$ is computed as:

\begin{equation}
a_{final} = 
\begin{cases}
\text{LLM}_{verifier}(q,M), & \text{sufficient;} \\
\text{LLM}_{CoT}(q), & \text{otherwise.}
\end{cases}
\end{equation}


\section{Experiments}

In this section, we empirically investigate the following five Research Questions (RQ): \textbf{RQ1}: Does our proposed DoM outperform baselines under KG incompleteness? \textbf{RQ2}: How robust is DoM when facing varying levels of KG incompleteness? \textbf{RQ3}: How does DoM perform when instantiated with different LLM backbones? \textbf{RQ4}: How does the core components contribute to the effectiveness of DoM? \textbf{RQ5}: How efficient is DoM in terms of token usage and runtime compared with baselines?

\subsection{Experimental Setup}

\subsubsection{Dataset} 
We evaluate DoM on the following datasets:
\begin{itemize}
    \item A collection of datasets proposed by Xu et al.~\cite{xu2024generate}, consisting of 1,000 sampled questions from each of the CWQ and WebQSP. For each dataset, four degrees of incomplete KGs are generated by randomly removing 20\%, 40\%, 60\%, or 80\% crucial triples, which appear in the gold relation path. These variants, denoted as IKG-20\%/40\%/60\%/80\%, are used to assess model robustness under varying degrees of KG incompleteness.
    \item A new dataset, IKGWQ, constructed by updating entity knowledge with recent facts and generating questions grounded in information missing from the existing KG.
\end{itemize}

\subsubsection{Baseline}
We evaluate DoM against a comprehensive set of baselines spanning three major categories in KGQA: (1) prompt-based LLM methods, including IO prompt~\cite{brown2020language} and CoT prompt~\cite{wei2022chain}; (2) fine-tuned LLM methods, such as ChatKBQA~\cite{luo2024chatkbqa} and RoG~\cite{luo2024reasoning}; and (3) retrieval-augmented inference frameworks, including ToG~\cite{sun2023think}, PoG~\cite{tan2025paths}, and GoG~\cite{xu2024generate}.

\subsubsection{Evaluation Metrics} Following previous works~\cite{li2024chain, luo2024reasoning, luo2024chatkbqa, xu2024generate}, exact match accuracy (Hits@1) is used as evaluation metric for all datasets.
\footnotetext[1]{Fine-tuned LLM backbones: RoG (LLaMA2-Chat-7B); ChatKBQA (Llama-2-7B).}
\begin{table}[t]
\centering
\begin{tabularx}{\linewidth}{
    l|*{3}{>{\centering\arraybackslash}X}
}
\toprule
\textbf{Method} & \textbf{IKGWQ} &  \textbf{CWQ} &  \textbf{WebQSP} \\
\midrule
\multicolumn{4}{c}{\textit{w.o. Knowledge Graph (DeepSeek-v3)}} \\
\midrule
IO & 28.5 & 50.1 & 68.8 \\
CoT & 51.0 & 54.3 & 66.3 \\
\midrule
\multicolumn{4}{c}{\textit{w.t. Knowledge Graph / Fine-tuned\textsuperscript{1}}} \\
\midrule
RoG & --  & 54.2  & 78.2 \\
ChatKBQA & --  & 39.3  & 49.5 \\
\midrule
\multicolumn{4}{c}{\textit{w.t. Knowledge Graph / Not-Training (DeepSeek-v3)}} \\
\midrule
ToG & 46.5  & 52.0  & 70.2 \\
PoG & 27.5  & 55.9  & 78.0 \\
GoG & 49.5  & 60.4  & 78.1 \\
DoM (Ours) & \textbf{84.5}  & \textbf{62.0}  & \textbf{81.7} \\
\bottomrule
\end{tabularx}
\caption{Performance on IKGWQ and IKG-40\% versions of CWQ and WebQSP. All results are reported as Hits@1 (\%). }
\vspace{-1.5em}
\label{table:main_result}
\end{table}

\subsubsection{Parameter Settings} In our experiments, we adopt five LLMs as the backbones: DeepSeek-v3, Qwen-max, Qwen2.5-72B, GPT-3.5 and GPT-4o\textsuperscript{2}. 
\footnotetext[2]{
Model versions: DeepSeek-v3 (default release); Qwen-max (qwen-max-2025-01-25); Qwen2.5-72B (qwen2.5-72b-instruct); GPT-3.5 (gpt-3.5-turbo-0613); GPT-4o (gpt-4o-2024-08-06).
}
All LLMs are accessed via their official APIs. The maximum token length is set to 512 for each call. The temperature is set to 0.4 during the information retrieval and exploration stages, and reduced to 0 during final answer generation to ensure deterministic outputs. The maximum KG search depth per sub-question ($W$) is set to 3, and the maximum number of sub-question iterations ($I$) is set to 6. For RAG retrieval, chunk size is 500 tokens, and the retrieval top-$k$ is set to 3.



\subsection{Main Results (RQ1)}
Table~\ref{table:main_result} reports Hits@1 scores of DoM and various baselines across multiple IKGQA datasets. As shown, DoM consistently outperforms all baselines.

On IKGWQ, DoM outperforms all evaluated baselines, demonstrating the effectiveness of its multi-source integration. The core challenge of IKGWQ lies in its faithful simulation of real-world KG incompleteness, characterized by irregularity and unpredictability. The unpredictability of missing knowledge fundamentally limits the effectiveness of existing KGQA methods. In IKGWQ, gold relation paths are often incomplete in the KG due to missing crucial relations or entities, posing significant challenges for inference. Even GoG, specifically designed for IKGQA, suffers significant performance degradation under these conditions. Due to missing facts in the KG, the retrieved context may lack essential information. This can mislead the LLM into generating incorrect answers, occasionally underperforming simpler prompting strategies such as CoT. In contrast, DoM introduces external unstructured information and employs a debate mechanism to integrate it with structured KG evidence. Benefiting from this multi-source integration, DoM surpasses the strongest baseline by 70.7\% on IKGWQ.
\begin{figure}[t]
    \centering
    \includegraphics[width=\linewidth]{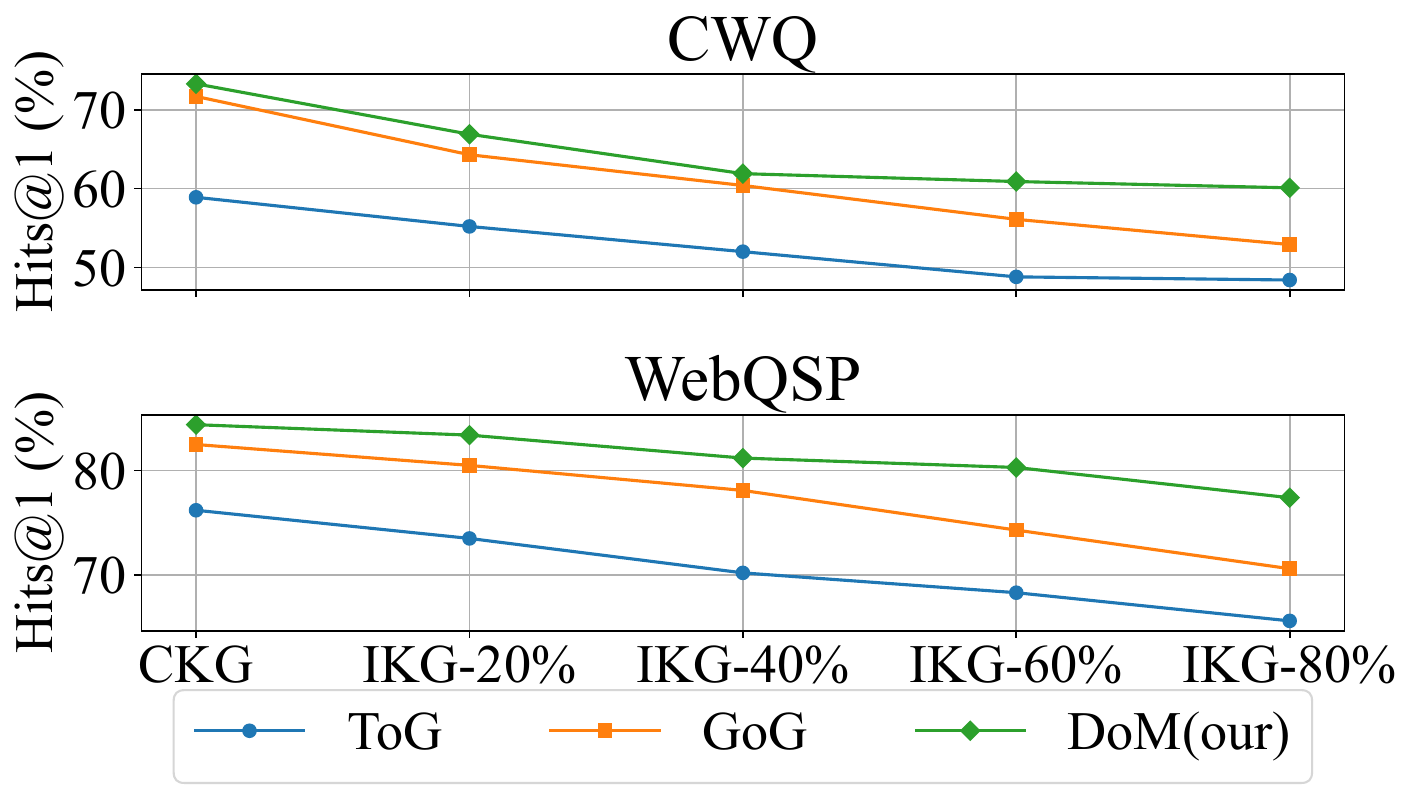}
    \caption{Performance on CWQ and WebQSP under varying KG incompleteness. CKG denotes a complete KG.}
    \label{fig:different_ikg}
\end{figure}

Beyond IKGWQ, DoM also achieves state-of-the-art performance on existing IKGQA datasets. While RoG outperforms several retrieval-based methods (e.g., GoG with DeepSeek-v3) on WebQSP, this primarily highlights that the performance of LLM-based inference methods depends heavily on the strength of the underlying backbone. In contrast, DoM consistently outperforms these baselines with the same backbone, demonstrating greater robustness and reasoning capability in IKG settings.


\subsection{Performance Under Varying KG Incompleteness (RQ2)}
To evaluate the robustness of DoM under different degrees of KG incompleteness, we conduct experiments on CWQ and WebQSP using DeepSeek-v3 as the backbone. The results, shown in Figure~\ref{fig:different_ikg}, demonstrate that DoM maintains consistent superiority across all degrees of incompleteness.

While all methods suffer performance degradation as more triples are removed, DoM exhibits notably slower decline, indicating enhanced resilience to missing knowledge. Even under extreme conditions (e.g., 80\% of critical triples removed), DoM still outperforms GoG by a large margin (with a 13.6\% relative improvement on CWQ IKG-80\%), highlighting its robustness to severe knowledge sparsity.

The improved stability of DoM stems from its ability to integrate heterogeneous evidence sources through structured agent interaction. In particular, the KG-based and RAG-based agents independently retrieve and infer from different modalities, while the Judge Agent coordinates their outputs. This architecture enables DoM to recover from incomplete retrieval and supports reliable multi-hop inference even under sparse KG conditions.

\subsection{Performance with Different Backbones (RQ3)}
\begin{table}[t]
\centering
\begin{tabularx}{\linewidth}{
    l|
    >{\centering\arraybackslash}X
    >{\centering\arraybackslash}X
    >{\centering\arraybackslash}X
}
\toprule
\multirow{2}{*}{\textbf{Backbone}} & \multicolumn{3}{c}{\textbf{IKGWQ}} \\
\cmidrule(lr){2-4}
 & \multicolumn{2}{c}{IKG} & NKG\\
\midrule
DeepSeek-v3     & \multicolumn{2}{c}{\textbf{84.5}} & 51.0 \\
Qwen-Max    & \multicolumn{2}{c}{78.0} & \textbf{52.5} \\
Qwen2.5-72B    & \multicolumn{2}{c}{69.0} & 35.5 \\
\midrule
\multirow{2}{*}{\phantom{\textbf{Method}}} & \multicolumn{3}{c}{\textbf{CWQ}} \\
\cmidrule(lr){2-4}
 & CKG & IKG-40\%  & NKG\\
\midrule
DeepSeek-v3     & \textbf{73.3}   & \textbf{62.0} & \textbf{54.3} \\
Qwen-Max     & 67.2   & 59.7 & 51.2 \\
Qwen2.5-72B     & 62.4   & 54.9 & 47.7 \\
\bottomrule
\end{tabularx}
\caption{Performance of DoM with different backbone models on IKGWQ and CWQ. CKG denotes a complete KG, and NKG corresponds to CoT reasoning without KG access.}
\vspace{-0.5em}
\label{table:different_backbones}
\end{table}

We evaluate how DoM performs with different LLM backbones to assess its adaptability to various models. As shown in Table~\ref{table:different_backbones}, DeepSeek-v3, which has the largest parameter size and strongest empirical performance among the evaluated models, achieves the best results on both IKG datasets. The performance gap between Qwen-Max and Qwen2.5-72B further shows that DoM benefits significantly from stronger backbones. This indicates that DoM can effectively scale with the capabilities of the underlying LLM.

\subsection{Ablation Study (RQ4)}
\begin{figure}[t]
    \centering
    \includegraphics[width=\linewidth]{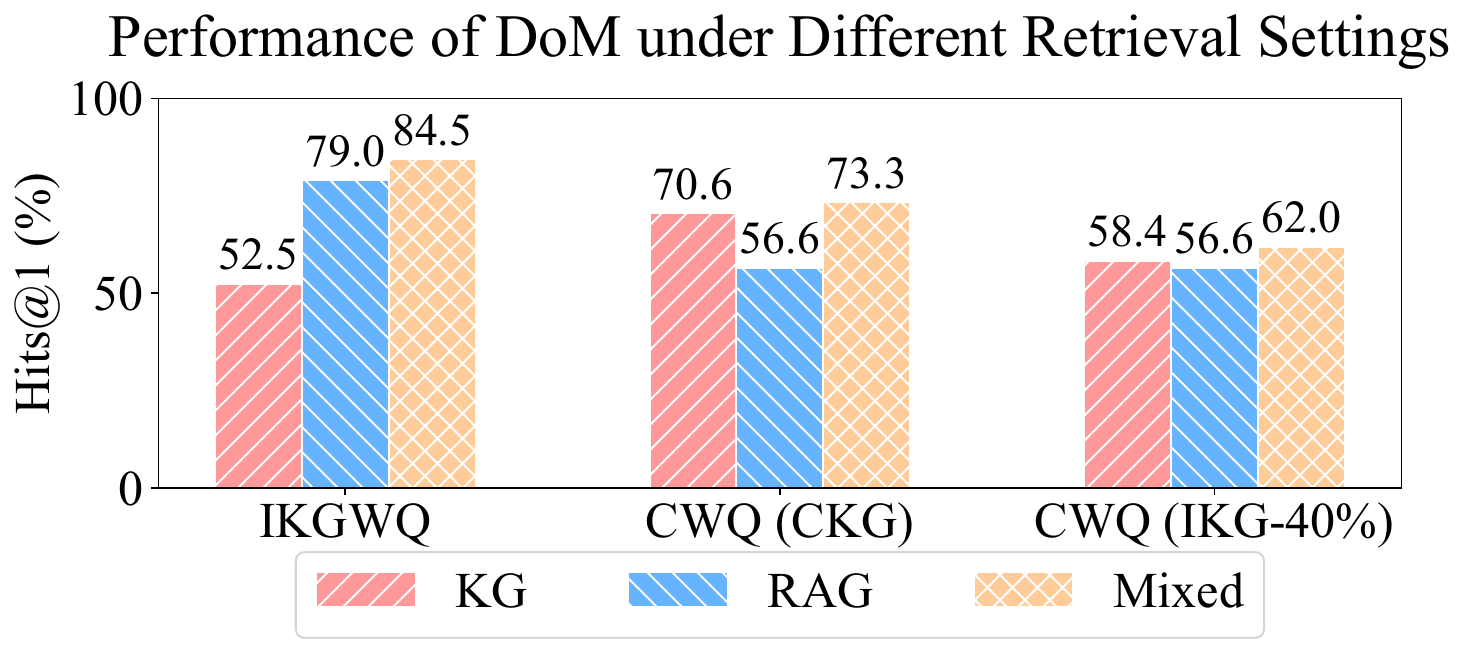}
    \caption{Performance of DoM with different retrieval agents on IKGWQ and CWQ. KG-retriever and RAG-retriever involve only the respective retrieval agent, with the Judge Agent reduced to a simple planner. Mixed-retriever activates all agents for full collaboration.}
    \label{fig:different_retriever}
\end{figure}


To assess the contribution of different retrieval agents in DoM, we conduct an ablation study using DeepSeek-v3 as the backbone, as shown in Figure~\ref{fig:different_retriever}. On the IKGWQ dataset, the KG Agent alone exhibits limited effectiveness, yielding performance comparable to that of LLM-based baselines reported in Table~\ref{table:main_result}. This can be attributed to the irregular and unpredictable missing patterns in IKGWQ, which hinder the KG Agent from consistently retrieving the necessary facts for accurate inference. In contrast, the KG Agent performs better on the CWQ dataset, where incompleteness is introduced in a controlled and predictable manner, making it easier to retrieve relevant facts. While the single-agent variants (KG-only and RAG-only) achieve reasonable performance individually, integrating their outputs via the Judge Agent yields significantly better results. These results underscore the complementary nature of structured and unstructured retrieval and validate the effectiveness of our modular design.
\begin{table}[t]
\begin{tabularx}{\linewidth}{
    l|*{4}{>{\centering\arraybackslash}X}
}
\toprule
            & ToG & PoG & GoG & DoM \\ 
\midrule
token / k   & 613 & 693 & 645 & 712       \\ 
time / min & 45  & 232 & 52  & 61        \\ 
\bottomrule
\end{tabularx}
\caption{Inference cost in terms of token usage and runtime across different methods.}
\vspace{-0.5em}
\label{table:cost}
\end{table}
It is notable that the KG-only variant of DoM outperforms the GoG baseline on IKGWQ. This can be attributed to the inherent nature of IKGWQ, where the missing information is more complex and unpredictable compared to the variants of CWQ and WebQSP. GoG relies heavily on retrieving relevant triples to complete the knowledge graph when missing facts need to be inferred. However, in IKGWQ, there often exists a significant gap between the retrieved triples and the missing facts required to answer the question, making it difficult for GoG to complete the inference process. In contrast, DoM's KG Agent can resort to LLM-based CoT inference when retrieval fails, leveraging LLM's internal knowledge to answer sub-questions. This adaptability allows DoM to handle the unpredictable incompleteness in IKGWQ more robustly, which explains its superior performance compared to GoG. This ablation confirms the necessity of both retrieval agents and the central role of the Judge Agent in coordinating complementary sources.

\subsection{Computational Cost Analysis (RQ5)}

To evaluate the computational overhead introduced by DoM, we analyze the inference cost on 100 samples from the CWQ IKG-40\% setting. As shown in Table~\ref{table:cost}, DoM introduces a moderate increase in token usage and runtime compared with GoG, primarily due to the additional retrieval and debate steps involving external evidence. Nevertheless, this overhead remains substantially lower than that of PoG, while yielding significantly better performance. Overall, although the integration of external textual knowledge inevitably adds some cost, the increase is modest and well compensated by the notable improvements in accuracy and robustness. These results demonstrate that DoM achieves an effective balance between computational efficiency and performance gains.



\section{Related Work}

\subsection{LLM-based KGQA}
To enhance faithfulness, recent advances have explored how LLMs can be combined with KGs in question answering~\cite{liu2020reasoning, huang2024joint}. These methods generally follow two main approaches: knowledge-internalization, where KGs are embedded into LLMs via fine-tuning to improve factual reasoning~\cite{li2023trea}, and knowledge-interaction, where LLMs query and perform inference on over external KGs~\cite{sun2023think}.

Despite their strong performance, these methods often rely on complete KGs, which are difficult to realize due to the high cost of constructing and maintaining KGs~\cite{hur2021survey}. This motivates growing interest in IKGQA~\cite{min2013distant, pflueger2022gnnq}. Existing IKGQA approaches can be broadly categorized into two classes. The first class focuses on KG-internal completion by predicting missing links based on relational patterns among existing triples~\cite{saxena2022sequence, zan2022complex, guo2023knowledge}. However, such methods exhibit limited effectiveness when dealing with newly emerged or out-of-KG knowledge. The second class addresses KG incompleteness by incorporating external information beyond the KG itself, e.g., retrieving unstructured textual corpora to provide supplementary evidence or construct question-specific subgraphs~\cite{sun2019pullnet, lv2020graph}, or utilizing LLMs as auxiliary knowledge sources~\cite{xu2024generate}.

\subsection{Multi-Agent Debate}
Multi-Agent Debate (MAD) enhances the reliability and diversity of LLM outputs by coordinating multiple agents under a judge's supervision, outperforming direct prompting on complex tasks~\cite{liang2023encouraging,chan2023chateval,qianscaling}.
Existing MAD systems typically follow two paradigms: adversarial debate, where agents present conflicting viewpoints to promote critical reasoning~\cite{liang2023encouraging,liu2024groupdebate}; and collaborative planning, where specialized agents cooperate to solve complex problems~\cite{du2023improving,haase2025beyond}.

Motivated by MAD's success, recent studies apply it to question answering~\cite{mao2025multi,ma2025debate,hu2024multi}. For example, DoG employs a MAD framework where agents reason over KG subgraphs for multi-hop questions~\cite{ma2025debate}. However, these works focus on inference and generation quality, with limited attention to addressing KG incompleteness.

\section{Conclusion}

To enable more effective inference under IKG scenarios, we proposed DoM, a MAD framework for IKGQA. By coordinating structured and unstructured evidence through agent collaboration, DoM effectively leverages their complementary strengths. Additionally, we constructed the IKGWQ dataset by revisiting samples from CWQ and WebQSP. We rebuild the corresponding QA pairs using up-to-date knowledge retrieved from reliable sources. This provides a more realistic benchmark for evaluating IKGQA systems. Extensive experiments demonstrate that DoM consistently outperforms strong baselines across multiple settings. We released both the IKGWQ dataset and our code to support future research.\footnote[3]{\url{https://github.com/liujilong0116/DoM}}

\section{Acknowledgments}
This work was supported by the National Key Research and Development Program under Grant 2023YFC2506800, and by the the National Natural Science Foundation of China under Grant No. 62406096.

\bibliography{aaai2026}



\clearpage

\section{Appendix}

\subsection{A. Algorithm of DoM}
\begin{algorithm}[h]
\caption{Debate over Mixed-knowledge (DoM) Framework}
\label{alg:dom_overview}
\begin{algorithmic}[1]
\REQUIRE Question $q$, Knowledge Graph $G = (E, R, T)$, External knowledge $\mathcal{R}$, Max steps $I$

\ENSURE Final answer $a_{final}$

\STATE $Q \leftarrow$ $\text{LLM}_{plan}(q)$ \COMMENT{Decompose into sub-questions}

\STATE $M \leftarrow Q$, \text{parameters} \COMMENT{Initialize system memory}
\STATE $i \leftarrow 0$
\WHILE{$i < I$ and $i < \text{len}(Q)$}
    \STATE $q_i \leftarrow Q[i]$ \COMMENT{Current sub-question}
    \STATE $\hat{a}_i^{\text{kg}} \leftarrow$ \textsc{KGAgent}$(q_i, G, M)$
    \STATE $\hat{a}_i^{\text{rag}} \leftarrow$ \textsc{RAGAgent}$(q_i, \mathcal{R}, M)$
    \STATE $a_i, q_{i+1}^{new} \leftarrow$ \textsc{JudgeAgent}$(\hat{a}_i^{\text{kg}}, \hat{a}_i^{\text{rag}}, M)$
    \STATE Update $Q$ with $q_{i+1}^{new}$
    \STATE Update $\hat{e}_{i+1}$ with $a_i$
    \STATE Update $M$ with $\hat{a}_i^{\text{kg}}$, $\hat{a}_i^{\text{rag}},$ $a_i$, Q
    \STATE $i \leftarrow i + 1$
\ENDWHILE

\STATE $a_{final} \leftarrow$ $\text{LLM}_{verifier}(q, M)$
\RETURN $a_{final}$



\end{algorithmic}
\end{algorithm}

\subsection{B. Baseline Introduction}

\subsubsection{Prompt-based LLM Methods} \

\textbf{IO Prompt}~\cite{brown2020language}: Directly prompts the LLM to generate answers in a single step without step-by-step inference. We follow the code released by ToG~\cite{sun2023think}.

\textbf{CoT Prompt}~\cite{wei2022chain}: Enhances IO prompting with chain-of-thought reasoning before producing the final answer. We follow the code released by ToG~\cite{sun2023think}.

\subsubsection{Fine-tuned LLM-based Methods} \
~\begin{figure*}[t]
    \centering
    \includegraphics[width=\linewidth]{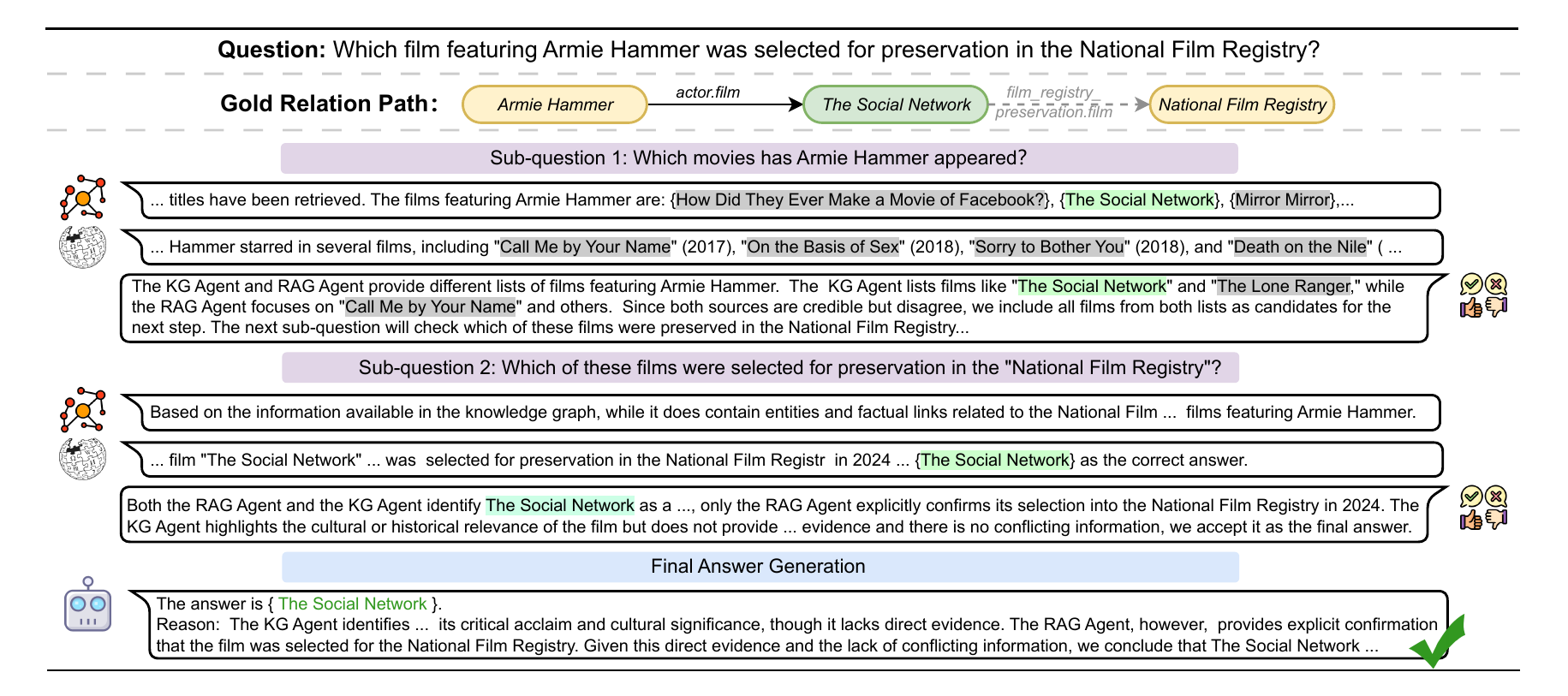}
    \caption{Case study. Example from the IKGWQ dataset illustrating how DoM's multi-agent debate framework handles KG incompleteness through iterative inference and evidence integration.}
    \label{fig:case_study}
\end{figure*}

\textbf{ChatKBQA}~\cite{luo2024chatkbqa}: A semantic parsing-based KGQA model that fine-tunes an LLM to generate SPARQL queries from natural language questions. We directly use the evaluation results provided in GoG~\cite{xu2024generate}.

\textbf{RoG}~\cite{luo2024reasoning}: A retrieval-guided method that fine-tunes LLMs to perform step-by-step inference along pre-retrieved KG paths, aligning outputs with KG structure to improve factual accuracy. We directly use the evaluation results provided in GoG~\cite{xu2024generate}.

\subsubsection{Retrieval-Augmented Inference Frameworks} \


\textbf{ToG}~\cite{sun2023think}: A training-free method using LLMs to perform multi-hop inference over KGs via beam search, with support for explicit path tracing and correction.


\textbf{PoG}~\cite{tan2025paths}: A retrieval-augmented method for multi-hop and multi-entity KGQA that improves faithfulness by dynamically pruning irrelevant paths using graph structure, LLM prompts, and pretrained encoders.


\textbf{GoG}~\cite{xu2024generate}: A training-free IKGQA method that enhances answer generation by inferring missing triples via a thinking-searching-generating framework combining KG context and LLM knowledge.

\subsection{C. Comparison between DoM and GoG under GPT-series models}

\begin{table}[h]
\centering
\begin{tabularx}{\linewidth}{
    l|*{3}{>{\centering\arraybackslash}X}
}
\toprule
\textbf{Method} & \textbf{IKGWQ} &  \textbf{CWQ} &  \textbf{WebQSP} \\
\midrule
\multicolumn{4}{c}{\textit{DeepSeek-v3}} \\
\midrule
CoT & 46.0 & 55.0 & 69.0 \\
GoG & 47.0  & 60.0  & 77.0 \\
DoM (Ours) & \textbf{82.0}  & \textbf{61.0}  & \textbf{80.0} \\
\midrule
\multicolumn{4}{c}{\textit{GPT-3.5}} \\
\midrule
CoT & 41.0 & 51.0 & 62.0 \\
GoG & 35.0  & 44.0  & 63.0\\
DoM (Ours) & \textbf{62.0}  & \textbf{56.0}  & \textbf{73.0} \\
\midrule
\multicolumn{4}{c}{\textit{GPT-4o}} \\
\midrule
CoT & 58.0 & 58.0 & 77.0 \\
GoG & 45.0  & 62.0  & 73.0 \\
DoM (Ours) & \textbf{79.0}  & \textbf{63.0}  & \textbf{83.0} \\
\bottomrule
\end{tabularx}
\caption{Performance of different methods with GPT-series models on 100-sample subsets of IKGWQ, CWQ (IKG-40\%), and WebQSP (IKG-40\%)}
\vspace{-1.2em}
\label{table:gpt_4o_result}
\end{table}



Table~\ref{table:gpt_4o_result} presents a detailed comparison between DoM and GoG across three backbone models: DeepSeek-v3, GPT-3.5, and GPT-4o, evaluated on IKGWQ, CWQ (IKG-40\%), and WebQSP (IKG-40\%). From the results, we observe that DoM consistently achieves the best performance across all backbone models. Both GPT-4o and DeepSeek-v3 outperform GPT-3.5 across datasets, indicating their stronger reasoning capacity and knowledge coverage. However, an intriguing observation is that on the IKGWQ dataset, GPT-4o underperforms DeepSeek-v3 under both GoG and DoM frameworks, despite showing superior performance on CWQ and WebQSP. 

We attribute this phenomenon to the nature of IKGWQ, which is designed to reflect more realistic and challenging IKG scenarios. Unlike CWQ and WebQSP, which simulate KG incompleteness through predefined deletions that result in more predictable missing facts, IKGWQ introduces more diverse and up-to-date facts that are not aligned with existing KG structures, leading to higher uncertainty and more complex inference demands.

While GPT-4o possesses powerful parametric knowledge, its strong generative capacity can sometimes introduce greater autonomy in reasoning trajectories, which may hinder performance in modular pipelines like GoG and DoM that rely on strict agent coordination and predictable intermediate steps. This issue is particularly evident in IKGWQ, where KG incompleteness is more complex and unpredictable, requiring consistent multi-turn interaction between agents. In contrast, CWQ and WebQSP adopt more patternized and constrained forms of incompleteness, where the inference paths tend to follow simpler and more regular structures. In such cases, GPT-4o's capabilities can be more effectively harnessed, leading to better performance. Meanwhile, DeepSeek-v3, though potentially less capable in terms of knowledge coverage, exhibits stronger consistency and alignment with structured multi-step inference workflows, making it more robust in IKGWQ.


\subsection{D. Manual Analysis}
\begin{figure}[t]
    \centering
    \includegraphics[width=\linewidth]{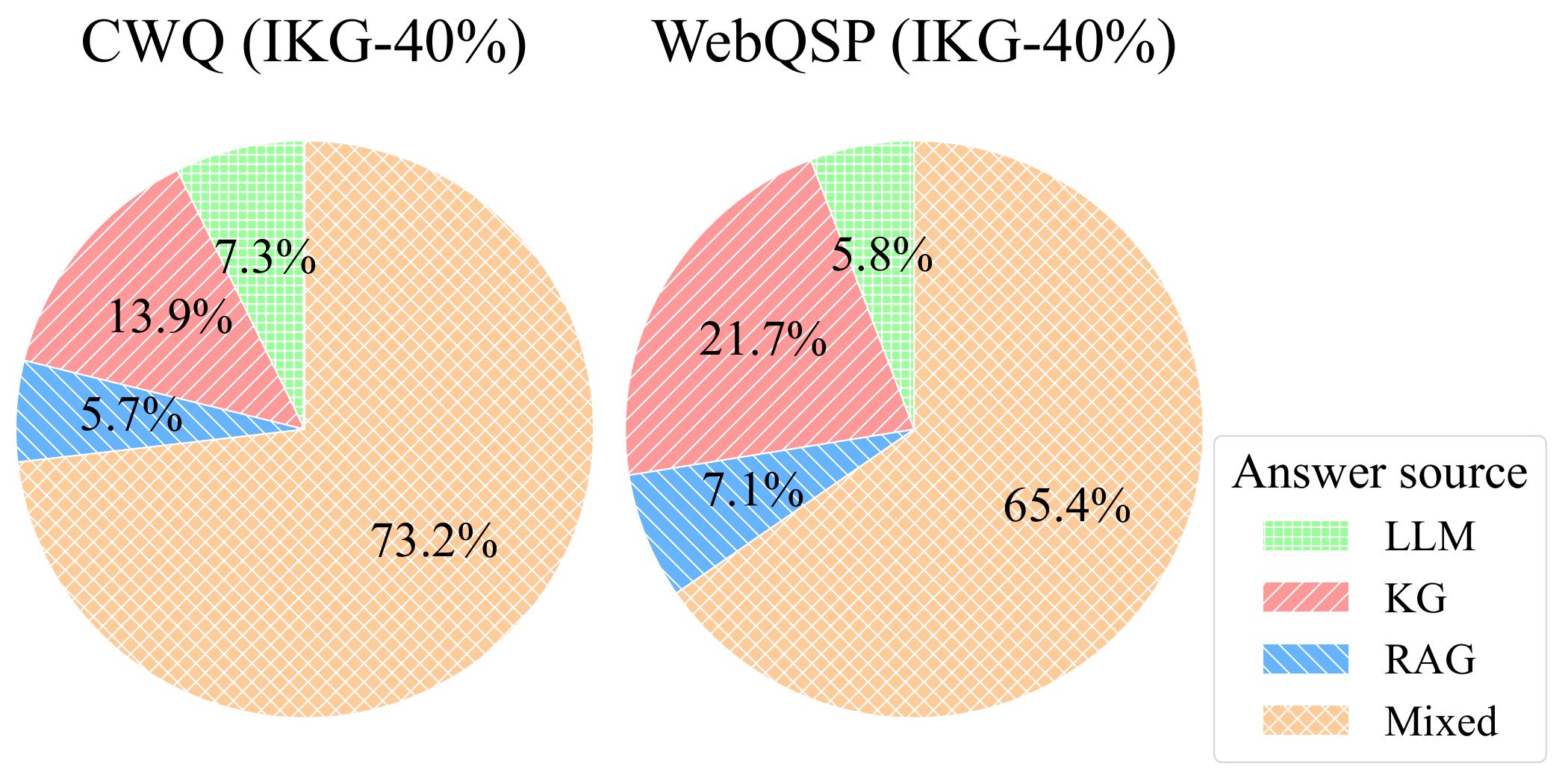}
    \caption{Answer source distribution for correctly answered samples in CWQ (IKG-40\%) and WebQSP (IKG-40\%).}
    \label{fig:manual_analysis}
\end{figure}

To examine how DoM leverages both structured and unstructured knowledge sources, we manually analyzed correctly answered samples from CWQ and WebQSP under the IKG-40\% setting. As shown in Figure~\ref{fig:manual_analysis}, a small portion of correct answers were derived from the LLM’s internal knowledge when neither the KG nor external data provided sufficient evidence. Most correct predictions, however, involved complementary evidence from both the KG and RAG agents, highlighting the value of dual-path retrieval. Through the Multi-Agent Debate (MAD) paradigm, our method enables interactive inference across heterogeneous knowledge sources, allowing agents to compensate for each other’s limitations. 

To further illustrate how multi-agent debate enables complementary inference under KG incompleteness, we present a representative example from the IKGWQ dataset (Figure~\ref{fig:case_study}). The question is: \emph{Which film featuring Armie Hammer was selected for preservation in the National Film Registry?}

In this example, the critical triple \emph{(The Social Network, film\_registry\_preservation.film, National Film Registry)} is missing from the KG. During the first iteration, the KG Agent successfully identifies \emph{The Social Network} as a relevant film based on structured KG facts, while the RAG Agent fails to retrieve useful evidence from external data. In the second iteration, the KG Agent can no longer offer useful knowledge due to the missing triple, but the RAG Agent locates supporting information in retrieved text. Through iterative inference and integration by the Judge Agent, the system ultimately arrives at the correct answer.

This case highlights how DoM enables complementary inference across structured and unstructured knowledge. When one agent is hindered by missing information, the other compensates, enabling robust inference for IKGQA.

\end{document}